\documentclass[runningheads]{llncs}
\usepackage{graphicx}
\usepackage[T1]{fontenc}    
\usepackage{hyperref}       
\usepackage{url}            
\usepackage{booktabs}       
\usepackage{paralist}       
\usepackage{amsfonts}       
\usepackage{nicefrac}       
\usepackage{microtype}      
\usepackage{xcolor}         

\usepackage{nameref}
\usepackage{wrapfig}
\usepackage{graphicx, float}
\usepackage{tikz}
\usepackage{soul}
\usepackage[inline]{enumitem}
\usepackage{amsmath,amsfonts,amssymb}
\usepackage{empheq,boldline,textcomp,gensymb, lipsum}
\usepackage{caption}
\usepackage{subcaption}
\usepackage{multirow}
\usepackage{array}
\newcolumntype{P}[1]{>{\centering\arraybackslash}p{#1}}
\newcolumntype{M}[1]{>{\centering\arraybackslash}m{#1}}
\usepackage[font=small,labelfont=bf]{caption}

\newcommand{\betti}[1]{\ensuremath{\beta_{#1}}}
\DeclareMathOperator{\im}{im}

\begin{document}
\title{Exploring the Geometry and Topology of Neural Network Loss Landscapes}
%
%
\author{Stefan Horoi\inst{\star,1,2}\orcidID{0000-0003-2951-2600} \and
Jessie Huang\inst{\star,3}\orcidID{0000-0002-5297-3563} \and
Bastian Rieck\inst{4}\orcidID{0000-0003-4335-0302} \and
Guillaume Lajoie\inst{1,2}\orcidID{0000-0003-2730-7291} \and
Guy Wolf\inst{\dagger,1,2}\orcidID{0000-0002-6740-059X} \and
Smita Krishnaswamy\inst{\dagger,3,5,6} \orcidID{0000-0001-5823-1985}}
\authorrunning{S. Horoi et al.}
%
\institute{Dept. of Math. and Stat., Université de Montréal, Montréal, QC, Canada  \and
Mila - Quebec Artificial Intelligence Institute, Montréal, QC, Canada \and
Dept. of Computer Science, Yale University, New Haven, CT, USA \and
Dept. of Genetics, Yale University, New Haven, CT, USA \and
Inst. of AI for Health, Helmholtz Centre Munich, Munich, Germany \and
\textbf{Corresponding author:} \email{
smita.krishnaswamy@yale.edu}
\email{}}

%
\maketitle              

\vspace{-10pt}
\begin{abstract}
Recent work has established clear links between the generalization performance of trained neural networks and the geometry of their loss landscape near the local minima to which they converge. This suggests that qualitative and quantitative examination of the loss landscape geometry could yield insights about neural network generalization performance during training. To this end, researchers have proposed visualizing the loss landscape through the use of simple dimensionality reduction techniques. However, such visualization methods have been limited by their linear nature and only capture features in one or two dimensions, thus restricting sampling of the loss landscape to lines or planes. Here, we expand and improve upon these in three ways. First, we present a novel ``jump and retrain'' procedure for sampling relevant portions of the loss landscape. We show that the resulting sampled data holds more meaningful information about the network's ability to generalize. Next, we show that non-linear dimensionality reduction of the jump and retrain trajectories via PHATE, a trajectory and manifold-preserving method, allows us to visualize differences between networks that are generalizing well vs poorly. Finally, we combine PHATE trajectories with a computational homology characterization to quantify trajectory differences.


\let\thefootnote\relax\footnote{\hspace{-4.3pt}$^\star$Equal contribution. $^\dagger$Equal senior-author contribution.}
\addtocounter{footnote}{-1}\let\thefootnote\svthefootnote

\let\thefootnote\relax\footnote{
This work was partially funded by Canada CIFAR AI Chair [G.L., G.W.], NSERC Discovery Grant RGPIN-2018-04821 [G.L.], Samsung Research Support [G.L.], NSERC CGSM and FRQNT B1X scholarships [S.H.]. 
The content is solely the responsibility of the authors and does not necessarily represent the views of the funding agencies.}
\addtocounter{footnote}{-1}\let\thefootnote\svthefootnote

\vspace{-20pt}
\keywords{Artificial neural network loss landscape  \and Non-linear dimensionality reduction \and Topological data analysis.}
\end{abstract}

\vspace{-20pt}
\section{Introduction}
\vspace{-7pt}

Artificial neural networks (ANNs) have been successfully used to solve a number of complex tasks in a diverse array of domains.
Despite being highly overparameterized for the tasks they solve, and having the capacity to memorize the entire training data, ANNs tend to generalize to unseen data. This is a spectacular feat since their highly non-convex optimization landscape should (theoretically) be a significant obstacle to using these models \cite{np_complete}. Questions such as why ANNs favor generalization over memorization and why they find ``good'' minima even with intricate loss functions still remain largely unanswered. One promising research direction is to study the geometry of the loss landscape of ANNs. Recent work tried to approach this task by proposing various sampling procedures and linear methods (based on PCA for example) for visualizing loss landscapes and their level curves. In some cases, this approach proved effective in uncovering underlying structures in the loss-landscape and linking them to network characteristics, such as generalization capabilities or structural features \cite{goodfellow2014qualitatively,keskar2016large_batch,im2016empirical_loss_surfaces,Goldstein_LL_NIPS2018_7875}. However, these methods have two major drawbacks: \textbf{(1)} they only choose directions that are linear combinations of parameter axes while the loss landscape itself is highly nonlinear, and \textbf{(2)} they choose only one or two among thousands (if not millions) of axes to sample and visualize while ignoring all others. 

First, an emerging challenge is how to sample and study such an extremely high dimensional optimization landscape (linear in the number of network parameters) with respect to minimized loss. We posit that one can utilize a manifold structure inherent to relevant connected patches of the loss landscape that are reachable during training processes in order to faithfully visualize the essential characteristics of its ``shape''. For this, we propose the {\em jump and retrain} method for sampling trajectories on the low loss manifolds surrounding found minima. The sampled points preserve information pertaining to the generalization capability of the neural network, while maintaining tractability of the visualization. 

We then utilize and adapt the PHATE dimensionality reduction method~\cite{phate}, which relies on diffusion-based manifold learning, to visualize these trajectories in low dimensions. In general, visualizations like PHATE are specifically designed to retain and compress as much variability as possible into two dimensions, and thus provide an advantage over previous linear approaches. Our choice of using PHATE over other popular methods, such as tSNE~\cite{tsne} or UMAP \cite{umap}, is due to its ability to capture both global and local structures of data. In particular, PHATE adequately tracks the continuous training trajectories that are traversed during gradient descent, while other methods tend to shatter them, and thus allows for significantly better visualizations of the manifolds on which these trajectories lie.

Finally, we turn to topological data analysis (TDA) methods to quantify features of the jump and retrain trajectories,  and thus characterize the loss-landscape regions surrounding different optima that emerge in networks that generalize well vs poorly. Our approach provides a general view of relevant geometric and topological patterns that emerge in the high-dimensional parameter space, providing insights regarding the properties of ANN training and reflecting on their impact on the loss landscape.

\vspace{-5pt}
\paragraph{Contributions:}
We present the jump and retrain sampling procedure in Section \ref{ss:methods_sampling} and show that the resulting data holds more relevant information about network generalization capabilities than past sampling procedures in Section \ref{ss:results_sampling}. We propose a new loss-landscape visualization method based on a variation of PHATE, implemented with cosine distance in Section \ref{ss:feature extraction}. Our visualization method is, to our knowledge, different from all other proposed methods for loss-landscape visualization in that it is naturally nonlinear and captures data characteristics from all dimensions. In Section \ref{ss:results_visualization}, we show that our method uncovers key geometric patterns characterizing loss-landscape regions surrounding good and bad generalization optima, as well as memorization optima. Finally, we use topological data analysis to characterize the PHATE transformed sampled manifolds and to quantify the differences between them in Section \ref{ss:results_tda}. To our knowledge this is the first time that a combination of data geometry (via PHATE) and topology has been used to analyze the loss landscape of ANNs.

\vspace{-5pt}
\section{Preliminaries}\label{s:preliminaries}
\vspace{-7pt}

\subsection{PHATE dimensionality reduction \& visualization}\label{ss:phate}
\vspace{-5pt}
Given a data matrix $\mathbf{N}$, PHATE first computes the pairwise similarity matrix $\mathbf{A}$ (using a distance function $\phi$ and an $\alpha$-decaying kernel), then row-normalize $\mathbf{A}$ to obtain the diffusion operator $\mathbf{P}$, a row-stochastic Markov transition matrix where $\mathbf{P}_{i,j}$ denotes the probability of moving from the $i$-th to the $j$-th data point in one time step. One of the reasons PHATE excels at capturing global structures in data, especially high-dimensional trajectories and branches, is that it leverages the diffusion operator (also used to construct diffusion maps~\cite{COIFMAN20065_Diffusion_Maps}) by running the implicit Markov chain forward in time. This is accomplished by raising the matrix $\mathbf{P}$ to the power $t$, effectively taking $t$ random walk steps, where $t$ is selected automatically as the knee point of the Von Neumann Entropy of the diffusion operator. 
To enable dimensionality reduction while retaining diffusion geometry information from the operator, PHATE leverages {\em information geometry} to define a pairwise {\em potential distance} as an M-divergence $\mathbf{ID}_{i,j}=\|\log P_{i,:} - \log P_{j,:}\|_2$ between corresponding $t$-step diffusion probability distributions of the two points, which provides a {\em global context} to each data point. The resulting information distance matrix $\mathbf{ID}$ is finally embedded into a tractable low-dimensional (2D or 3D) space by metric multidimensional scaling (MDS), thereby squeezing the intrinsic geometric information to calculate the final 2D or 3D embeddings of the data. For further details, see Moon et al.~\cite{phate}.

\vspace{-5pt}
\subsection{Topological data analysis}\label{sec:Topological data analysis}
\vspace{-5pt}
Topological data analysis~(TDA) refers to a set of techniques for
understanding complex datasets by means of their topological features,
i.e., their connectivity~\cite{Edelsbrunner10}. While TDA is applicable
in multiple contexts, we focus specifically on the case of graphs. Here,
the simplest set of topological features is given by the number of
connected components~$\betti{0}$ and the number of cycles~$\betti{1}$,
respectively. Such counts, also known as the \emph{Betti numbers}, are
coarse graph descriptors that are invariant under graph isomorphisms.
Their expressivity is somewhat limited, but can be increased by
considering a function $f\colon V \to \mathbb{R}$ on the vertices of a
graph~$G = (V, E)$ with vertex set~$V$ and edge set~$E$. Since $V$ has
finite cardinality, so does its image~$\im f$, i.e.,
$\im f := \{w_1, w_2, \dots, w_n \}$.
Without loss of generality, we assume that $w_1 \leq \dots \leq w_n$. We
write $G_i$ for the subgraph induced by filtering according to $w_i$,
such that the vertices of $G_i$ satisfy $V_i := \{ v \in V \mid f(v)
\leq w_i\}$, and the edges satisfy $E_i := \{ (u, v) \in E \mid
\max(f(u), f(v)) \leq w_i \}$. The subgraphs $G_i$ satisfy a nesting
property, as $G_1 \subseteq G_2 \subseteq \dots \subseteq G_n$. It is
now possible to calculate topological features alongside this
\emph{filtration} of graphs, tracking their appearance and
disappearance. If a topological feature is created in $G_i$, but
destroyed in $G_j$~(it might be destroyed because two connected
components merge, for instance), we represent this by storing the point
$(w_i, w_j)$ in the \emph{persistence diagram} $\mathcal{D}_f$ associated to $G$.
Persistence diagrams are known to be salient descriptors of graphs and have seen increasing usage in graph classification~\cite{Hofer17,Hofer20,Rieck19b,Zhao19}. Their primary appeal lies in their capability to summarize shape information and the
robustness to noise~\cite{Cohen-Steiner07} of topological features made them successful shape descriptors in a variety of applications~\cite{Amezquita20,Rieck20}. Numerous fixed filtrations have been described for different tasks~\cite{Hofer20,Zhao19}, but in our context, a natural choice for~$f$ is provided by the \emph{loss function} of the network itself. This will enable us to describe the topology of the loss landscape.



\vspace{-5pt}
\section{What is the ``shape'' of the loss landscape?}\label{s:methods}
\vspace{-7pt}

The loss landscape of an ANN
can be formulated mathematically as the geometry and topology defined in the high dimensional parameter space $\Theta$ by a loss function $f\colon\Theta \to \mathbb{R}$ that assigns a loss value $f(\theta)$ to every possible parameter vector $\theta$ (e.g., consisting of network weights) based on considered training or validation data. While $f(\theta)$ provides some information for examining and filtering the various configurations of model parameters, the exceedingly high dimensionality of the parameter space (i.e., often in the millions) renders the task of visualizing or analyzing the entire loss-landscape over $\Theta$ virtually impossible. However, since the optimization process considered in this context is guided by the objective of minimizing the loss, we can expect most regions in the high-dimensional $\Theta$ to be of negligible importance, if not unreachable, for the network training dynamics or the viable configurations learned by them. Therefore, the analysis of the loss landscape can focus on regions that are reachable, or reliably traversed, during this optimization process, which we expect would have a much lower intrinsic dimensionality than the ambient dimensions of the entire parameter space. 

Our approach to characterizing the ``shape'' of the loss landscape in such local regions of interest is inspired by the construction of a tangent space of manifolds in intrinsic terms in Riemannian geometry.
There, tangent vectors at a given point of interest are defined by aggregating together intrinsic trajectories (on the manifold) traversing through the tangential point. This aggregation, in turn, yields equivalence classes that signify tangential directions, whose span is considered as the tangent space that provides a local intrinsic (typically low-dimensional) coordinate neighborhood in the vicinity of the tangential point. 
In a similar way, here we propose to leverage trajectories of the network optimization process in order to reveal the intrinsic geometry exposed by them as they flow towards convergence to (local) minima of the loss. The remainder of this section provides  a detailed derivation of the three main steps in our approach.

\vspace{-5pt}
\subsection{Jump and retrain sampling}\label{ss:methods_sampling}
\vspace{-5pt}
Gradient-based optimization methods naturally explore low-loss regions in parameter space before finding and settling at a minimum. We hypothesize that keeping track of optimizer trajectories in parameter space is an efficient way of sampling these low-loss manifolds and thus to gather information about the relevant part of the loss regions surrounding minima. This can be seen as an approximation to the Morse--Smale complex~\cite{Gyulassy08}, a decomposition of~$f$ into regions of similar gradient behavior, whose analytical calculation is infeasible given the overall size of $\Theta$. With this in mind we have designed the following ``jump and retrain'' or J\&R sampling procedure. Most of our experiments were conducted with \texttt{SEEDS}$=\{0,1,2,3\}$ or $\{0,1,2,3,4\}$, \texttt{STEP\_SIZES}$=\{0.25, 0.5, 0.75, 1.0\}$ and $N=40$ or $50$. Let $\theta_o$ represent the vector of network parameters at the minimum:

\begin{itemize}[nosep, leftmargin=*]
    \item For $seed\in$ \texttt{SEEDS}:
    \begin{itemize}[nosep]
        \item For $step\_size\in$ \texttt{STEP\_SIZES}: 
        \begin{enumerate}[nosep]
            \item Choose a random $v_{seed}$ in  $\Theta$ 
            and filter-normalize to obtain $\overline{v}_{seed}$;
            \item Set the ANN parameters 
            to be $\theta_{\text{jump-init}} = \theta_o + step\_size \cdot \overline{v}_{seed}$;
            \item Retrain for $N$ epochs with the original optimizer; 
            \item Record parameters $\theta$ and evaluate the loss at each retraining epoch;
        \end{enumerate}
    \end{itemize}
\end{itemize}


\vspace{-5pt}
\subsection{PHATE dimensionality reduction and visualization}\label{ss:feature extraction}
\begin{figure}[ht]
\vspace{-5pt}
    \centering
    \includegraphics[width=0.9\textwidth]{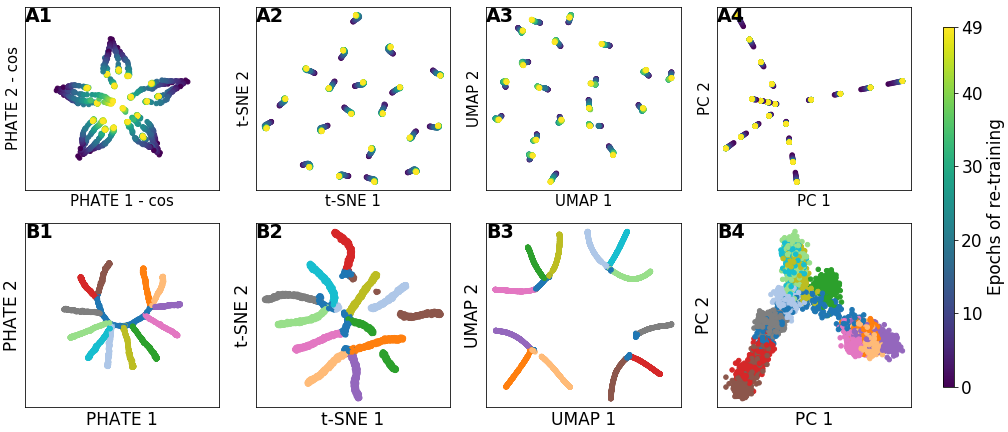}
    \caption{2D embeddings of the J\&R experiment (5 random seeds) results for a WResNet28-2 network using PHATE (\textbf{A1}), t-SNE (\textbf{A2}), UMAP (\textbf{A3}) and the linear method PCA (\textbf{A4}). \textbf{B:} embeddings of an artificial data set having a fully connected tree-like structure found using the same techniques. We see that PHATE consistently retains continuous trajectory structures while other embeddings (tSNE/UMAP) shatter the structure, or miss important features (PCA) because of uninformative projections to low dimensions.
    }
    \label{fig:dim_red_comparison}
    \vspace{-15pt}
\end{figure}

\vspace{-5pt}
Since the points sampled with the presented procedure are not simply positioned on a line or a plane, methods that project these points onto a 1D or 2D space are unable to properly visualize all the variation in the data.
PHATE has allowed us to bypass the key drawbacks of previously proposed linear visualization methods by \textbf{(1)} Capturing variance in the sampled data from all relevant dimensions and embedding it in a low-dimensional space; and \textbf{(2)} preserving high-dimensional trajectories and global structures of data in parameter space. All modern dimensionality-reduction techniques would have achieved \textbf{(1)} with varying degrees of success. However our proposed cosine-distance PHATE, which uses cosine distance to compute the pairwise similarity matrix $\mathbf{A}$ and to perform MDS, has unique advantages over other state-of the art dimensionality reduction methods to accomplish \textbf{(2)}. Figure~\ref{fig:dim_red_comparison} demonstrates this by showing a comparison of multiple such techniques, namely PHATE, PCA, t-SNE \cite{tsne} and UMAP \cite{umap}, and how they each embed the data from the J\&R sampling (Fig. \ref{fig:dim_red_comparison}\textbf{A})  and an artificial data set having a tree-like structure (Fig. \ref{fig:dim_red_comparison}\textbf{B}) in a 2D space.

While some trajectory-like structure is visible in all low-dimensional embeddings, only PHATE properly captures intra-trajectory variance. PHATE is also the only technique that captures the global relationships between trajectories while t-SNE and UMAP have a tendency to cluster points that are close in parameter space and disregard the global structure of the data. On the artificial data set (Fig. \ref{fig:dim_red_comparison} \textbf{C}), what we observe is that the embeddings of the linear method PCA are highly affected by the noise in the data while t-SNE and UMAP have a tendency of shattering trajectories that should be connected.  By accomplishing \textbf{(1)} and \textbf{(2)} PHATE effectively reconstructs the manifold in parameter space from which the jump and retrain data is sampled and allows its embedding in a lower-dimensional space preserving its local and global structure.

\vspace{-5pt}
\subsection{Topological feature extraction}\label{ss:quantifying}
\vspace{-5pt}
In order to quantify the shape of the PHATE embeddings, we calculate a set of topological features. To this end, we first compute a kNN graph~(with $k = 20$) based on the PHATE diffusion potential distances. Each node of this graph~$G$ corresponds to a specific point in parameter space~$\theta \in \Theta$ sampled from the loss landscape. We obtain a filtration function from this by assigning each vertex $\theta$ 
its corresponding loss value $f(\theta)$. Then we filter over this graph by slowly increasing the loss threshold $t$, which effectively reveals increasingly larger parts of the graph and creates a filtration as described in Section~\ref{sec:Topological data analysis}. From this filtration we obtain a set of persistence diagrams $\mathcal{D}_0$, $\mathcal{D}_1$ summarizing the topological features of the respective embedding~(we omit the index~$f$ for simplicity). 
%
As a powerful summary statistic, we calculate the \emph{total persistence}~\cite{Cohen-Steiner07} of a persistence diagram~$\mathcal{D}$, i.e.,
$\operatorname{pers}(\mathcal{D}) := \sum_{(c, d) \in \mathcal{D}} \left|d - c\right|^2$. $\operatorname{pers}(\cdot)$ serves as a complexity measure that enables us to compare different embeddings. This measure has the advantage that it is invariant with respect to rotations of the embedding. Moreover, it satisfies robustness properties, meaning that it will change continuously under a continuous perturbation of the input filtration.

\vspace{-5pt}
\section{Geometric and topological reflection on ANN training and generalization}\label{s:results}
\vspace{-7pt}
\subsection{Experimental setup}\label{ss:experimental_setup}
\vspace{-5pt}
In order to assess the effectiveness of our loss landscape visualizations and characterizations, we trained wide ResNets \cite{wide_resnet} of varied sizes with depth $\in \{10, 16, 22\}$, width $\in\{1,2\}$ on the CIFAR10 image classification task \cite{cifar10} from initialization to optimum. The networks were trained with a combination of the following hyperparameters: batch size $\in\{32, 128, 512\}$, weight decay $\in\{0, 0.0001, 0.001\}$ and either with or without data augmentation (random horizontal flips and crops). Each of the 108 WResNets were initialized identically and trained for 200 epochs using SGD with 0.9 momentum and a learning rate with initial value of 0.1 followed by step decay by a factor of 10 at epochs 100, 150 and 185.

Utilizing data augmentation and weight decay has allowed us to find ``good optima'' that generalize better, increasing test accuracy from around $\sim$85\% (for ``bad optima'' trained without d.a. and w.d.) to $\sim$95\%. 
We also trained a WResNet28-2 network for memorization by completely randomizing the labels of the training data. This modification was presented in \cite{Zhang2017UnderstandingDL} to show that neural networks have the ability to completely memorize the training data sets.


\vspace{-5pt}
\subsection{Jump and retrain sampling captures generalization and training characteristics}\label{ss:results_sampling}
\vspace{-5pt}

\begin{table}[b]
\centering
\vspace{-20pt}
\caption{Mean and standard error (\%) of the 11 classifiers accuracies on the 5 class generalization, weight decay and data aug. classification tasks with different features.}
\label{tab:sampling_classif}
 \begin{tabular}{|M{7cm}|M{1.5cm}|M{1.5cm}|M{1.5cm}|}
 \hline
 Features & 5 class gen. & Weight decay & Data augmentation \\
 \hline
 Theoretical random (1/\#classes) & 20.0 & 33.3 & 50.0\\
 \hline
 Randomized J\&R retrain loss and accuracy values & 20.7 $\pm$ 1.1 & 38.9 $\pm$ 1.5 & 52.7 $\pm$ 1.9 \\
 \hline \hline
 Grid sampling,  train loss and accuracy values & 30.1 $\pm$ 2.7 & 51.1 $\pm$ 2.9 & 62.3 $\pm$ 5.1 \\  
 \hline
 Naive sampling, train loss and accuracy values & 31.8 $\pm$ 4.0  & 55.7 $\pm$ 3.9 & 67.6 $\pm$ 5.2  \\
 \hline
 J\&R sampling, retrain loss and accuracy values & \textbf{39.2} $\pm$4.3  & \textbf{58.2} $\pm$3.6 &  \textbf{72.1} $\pm$5.4 \\
 \hline
\end{tabular}
\end{table}

Past work has shown that networks with different generalization capabilities tend to have optima surrounded by regions of distinct geometrical characteristics \cite{Hochreiter1997_flat_minima,im2016empirical_loss_surfaces,Goldstein_LL_NIPS2018_7875,keskar2016large_batch,entropy_sgd}. Regularization techniques, such as weight decay and data augmentation, are believed to play a role in this difference in geometries \cite{entropy_sgd,im2016empirical_loss_surfaces}. Inspired by past results, we formulate the following classification tasks to evaluate loss landscape sampling methods and see if the sampled data holds information about the networks ability to generalize
and the geometry of the loss landscape. We used the sampled loss and accuracy values as features and we separated the trained WResNet models into 5 (almost) equally-sized ``generalization'' classes according to the value of the test loss at optimum. The $\sim$20\% of networks with the lowest test losses at optimum were assigned to class 1, and so on. We then trained 11 simple classifiers 
to predict the generalization class of each network with training losses and accuracy features. All results were obtained from 10-fold cross validation. It is important to note that the classifiers were not tuned to favor any of the sampling procedures. Two similar classification tasks were designed to predict weight decay and whether or not data augmentation was used.

To evaluate the effectiveness of the jump and retrain sampling, we compared it to
two other sampling procedures. We refer to the first as {\it grid sampling}, and it is directly inspired by the 1D or 2D linear interpolations used in past visualization methods \cite{goodfellow2014qualitatively,im2016empirical_loss_surfaces,Goldstein_LL_NIPS2018_7875}. Here, we randomly choose 3 vectors starting from the optimum ($\theta_o$) and construct a 3D grid using the 3 vectors as basis. The loss and accuracy on the training set is then evaluated at all points on the grid. The second comparison sampling procedure, that we call {\it naive sampling}, evaluates the loss in random directions ($\theta_i$) centered at the optimum $\theta_0$, and multiple step sizes $c$; i.e. evaluate at $\theta_o + c\theta_i$. This method tests whether using more directions and step sizes when sampling, without the grid-like structure, is more informative since it explores a greater number of directions in parameter space. All methods considered, including J\&R, sample 640 points from the loss landscape excluding the optimum itself. 
We applied the filter-wise normalization presented in \cite{Goldstein_LL_NIPS2018_7875} when obtaining random directions.
As a control experiment, we trained the same classifiers on scrambled versions of the best performing features, making sure classifiers were not overfitting the data and evaluating the impact of feature distributions alone. The results are shown in Table \ref{tab:sampling_classif}.

Using the loss and accuracy values sampled with the J\&R procedure as features allows the classifiers to achieve mean accuracies of 39.2\%, 58.2\% and 72.1\% on the generalization, weight decay and data augmentation classifications tasks respectively. This is significantly higher than the mean classification accuracies reached using the data sampled with the other two methods. We confirmed the validity of the classification accuracy with J\&R data with permutation tests where the accuracy is essentially the same as random. Indeed, our results indicate that connected patches of the low-loss manifold surrounding the optima, which are found with the jump and retrain procedure, hold more information about the region's geometry and the network's ability to generalize at that optimum than the data sampled with non-dynamical methods. Furthermore, the success of the J\&R sampled data on the weight decay and data augmentation classification tasks shows that our sampling method captures information not only about generalization but also about the training procedure. The set of J\&R sampled loss and accuracy values seem to have distinct characteristics depending on the training procedure used to reach the minima and whether or not regularization methods were used and to what extent. This helps support the idea that a dynamical sampling of the loss landscape, which mimics the behavior of optimization procedures, is more informative than static sampling methods. 

\vspace{-7pt}
\subsection{Generalization indicated by visual patterns from loss landscape regions around optima}\label{ss:results_visualization}
\vspace{-7pt}

\begin{figure}[ht]
    \centering
    \includegraphics[width=0.9\textwidth]{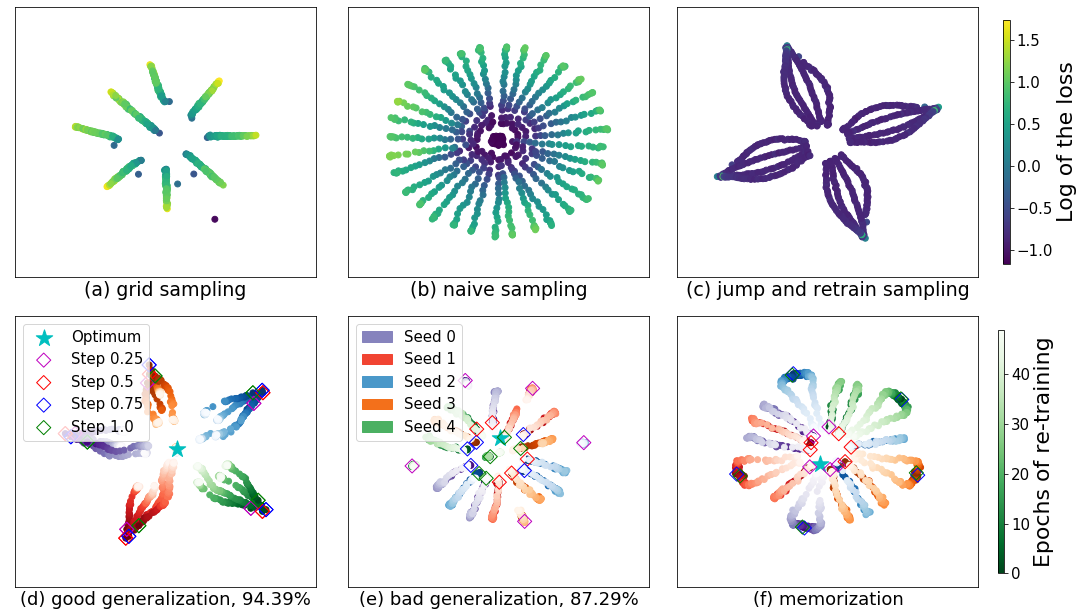}
    \caption{\textbf{Top row:} PHATE embeddings of the data generated with the three sampling procedures colored by loss in log scale. The rigidity of the grid and naive sampling methods makes visualizations less informative.
    The J\&R sampling successfully samples the low-loss manifolds surrounding minima.
    \textbf{Bottom row:} PHATE embeddings comparison between good, bad and memorization minumum.
    The $\theta_{\text{jump-init}}$ points
    are marked by diamonds of colors corresponding to $step\_size$ and trajectories are colored by $seed$
    with descending hue, i.e. the color gets whiter as retraining progresses. In contrast to more continuous trajectories returning to near the optimum in the good generalization case, bad generalization and memorization display more random patterns where weights move out before moving back, often switching direction during retraining. 
    }
    \label{fig:visualization}
    \vspace{-15pt}
\end{figure}

PHATE visualization of the data sampled with the J\&R procedure as in  Fig.~\ref{fig:visualization}(c) clearly demonstrate the trajectories and the low-loss manifold surrounding the found optima that was actually traversed during training. It is more informative that the visualization of data from the other sampling procedures (Fig~\ref{fig:visualization}(a,b)). Also, although all networks achieved $\sim0$ loss on their respective training sets and only a $\sim$7\% difference in their test set accuracies, Fig.~\ref{fig:visualization}(d,f) reveal stark differences between network configurations  that memorize (or overfit) versus ones that generalize. The good generalization minimum has a distinctive star shaped pattern. This indicates that even when points are thrown away from the minima they return to it immediately without traveling outward. Thus the minima seems to serve as an effective attractor to which trajectories repeatedly return.

In case of bad minimum, the trajectories start off near the middle of the plot (darker points in the middle of Fig.~\ref{fig:visualization}e) but, during the retraining, they diverge toward the edges before occasionally coming back towards the middle of the plot. This outward movement is in stark contrast with the consistent retraining trajectories surrounding the good generalization minimum, which immediately return to the valley. In this sense the minima are not stable and perturbations of the parameters cause networks to escape this minima.   The ``memorization'' minimum plot (Fig.~\ref{fig:visualization}f) looks similar to the bad generalization plot, with trajectories that go outward at small step sizes of the jump. However, curiously at larger step sizes, the trajectories seem to return without going outward first, but they do not return immediately, they show some lateral movement, potentially indicating bumpiness in the landscape that they are avoiding.


\begin{figure}[ht]
    \centering
    \includegraphics[width=0.9\textwidth]{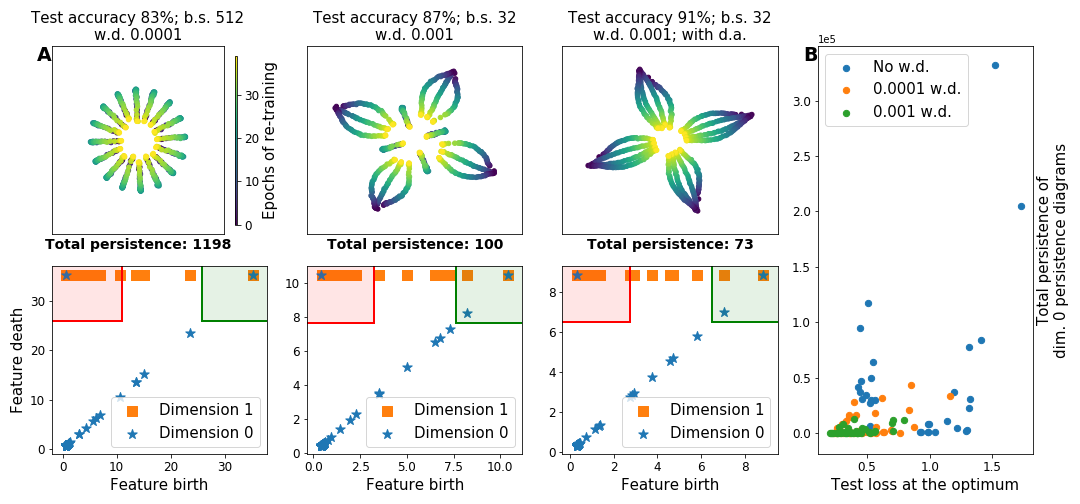}
    \caption{\textbf{A, top row:} 2D PHATE embeddings of the data sampled with the jump and retrain procedure surrounding minima reached by four WResNet16-2 with the same initialization but trained in different ways and thus reaching different accuracies on the test set. \textbf{A, bottom row:} Persistence diagrams of the loss-level filtration computed from the respective PHATE diffusion potentials. The total persistence associated with each optima/persistence diagram is written in bold. Both PHATE and the topological features of the loss landscape seem to differentiate these four networks which have very different generalization capabilities. \textbf{B:} Total persistence of dimension 0 computed from the jump \& retrain data for each network w.r.t.\ the test loss at optimum. Colors indicate the value of the weight decay used during training.}
    \label{fig:tda_characterization}
    \vspace{-15pt}
\end{figure}

\vspace{-5pt}
\subsection{Generalization may be related to low topological activity in near-optimum regions}\label{ss:results_tda}
\vspace{-5pt}
PHATE has allowed us to generate low dimensional representations of the low-loss manifolds sampled through the jump and retrain procedure. Here, topological data analysis enables us to quantify the topological features of these manifolds and thus characterize the loss landscape regions surrounding different optima. In Fig.~\ref{fig:tda_characterization} \textbf{A} we show the PHATE visualizations of the jump and retrain data sampled around different optima of a WResNet16-2 network and the corresponding dimension 0 and 1 persistence diagrams. From the PHATE visualizations alone it is clear that the sampled manifold have different structures. In fact, the persistence diagrams confirm that these manifolds also have different topological features. In particular, optima that generalize poorly seem to be surrounded by manifolds with more topological activity, the two persistence diagrams on the left having more high-persistence points (higher density of points in the red rectangles of Fig. \ref{fig:tda_characterization} \textbf{A}).
Conversely, networks that generalize well have both zero and one dimensional topological features emerging later in the loss threshold (green rectangles). 


In order to verify this observation in a more general case, we also computed the \emph{total persistence}~(see Section~\ref{ss:quantifying})
of the persistence diagrams corresponding to 
each one of the 108 trained WResNets. In Fig.~\ref{fig:tda_characterization} \textbf{B} we plot these values as functions of the test loss at optimum and color the points according to the value of weight decay used during training to reach those points. The first thing we observe is that optima surrounded by regions of high topological activity tend to have a higher loss value at the optima, while low-loss optima have a lower associated total persistence. This further confirms the idea that good generalization optima are surrounded by relatively flat loss-landscape regions while bad generalization optima tend to be situated in regions with many non-convexities. Furthermore we observe that optima found with the most aggressive weight decay (namely 0.001) are surrounded by regions of relatively low topological activity while optima reached without the use of weight decay are associated with the highest levels of total persistence. These results seem to suggest that the use of weight decay, an efficient regularization method, allows optimizers to find minima on low-loss manifolds with low-persistence topological features. Past results have linked the use of regularization techniques to finding good generalization minima surrounded by flat regions, i.e. regions of low geometrical activity. We have expanded on these results by showing that the topological activity in those regions is also relatively low when compared to regions surrounding optima found without regularization.

\vspace{-13pt}
\section{Discussion and conclusion}
\vspace{-10pt}
We propose a novel approach to dynamically sample the loss landscapes of deep learning models which takes theoretical inspiration from the fields of Riemannian geometry and dynamical systems. Our sampling method efficiently samples points form the low-loss manifolds surrounding minima found through gradient descent. The resulting sampled data holds more information than past loss landscape sampling methods about the geometry of the loss landscape, the network ability to generalize at the optimum and the training procedure and regularization used to reach that optimum. We then present a new loss landscape visualization method based on the state-of-the-art PHATE dimensionality reduction method, which is able to reconstruct the high-dimensional trajectories sampled in two dimensional representations. Our approach enables geometric exploration of the sampled manifolds and regions surrounding generalization and memorization optima, found via ANN training, to provide insight into generalization capabilities and training of the network. Finally, topological data analysis enables us to characterize these regions through the computation of their topological features. We found that weight decay, a powerful regularization technique, allows ANN optimizers to find minima in regions of lower topological activity. An interesting research direction would be to try to apply dimensionality reduction techniques that better take into account the time dependency of the data. We expect in future work our sampling, visualization and topological characterization approaches to enable more methodical paradigms for the development of ANNs that generalize better, train faster, and to provide fundamental understanding of their capabilities.

\bibliographystyle{splncs04}
\bibliography{main}

\end{document}